\documentclass{egpubl}
\usepackage{arxiv_conf}
\usepackage{amssymb}
\usepackage{mathtools}
\usepackage{amsmath}
\usepackage{array}
\usepackage{booktabs}
\usepackage{makecell}
\usepackage{bbm}
\usepackage{url}

\DeclareMathOperator*{\argmax}{arg\,max}

\DeclarePairedDelimiter\norm{\lVert}{\rVert}

\WsPaper

\usepackage[T1]{fontenc}
\usepackage{dfadobe}

\usepackage{cite}

\BibtexOrBiblatex

\electronicVersion
\PrintedOrElectronic

\ifpdf \usepackage[pdftex]{graphicx} \pdfcompresslevel=9
\else \usepackage[dvips]{graphicx} \fi

\usepackage{egweblnk}

\title{DesigNet: Learning to Draw Vector Graphics as Designers Do}

\author[Tomas Guija-Valiente \& Iago Suárez]
{\parbox{\textwidth}{\centering
        Tomas Guija-Valiente$^{1,2}$\orcid{0009-0000-0911-3317}
        and Iago Suárez$^{1,3}$\orcid{0000-0003-4006-4378}
        }
        \\
{\parbox{\textwidth}{\centering $^{1}$~Machine Learning Circle, Spain\\
         $^2$~Universidad Politécnica de Madrid, Departamento de Inteligencia Artificial, Spain \\
         $^3$~Qualcomm XR Labs, Spain
       }
}
}
\begin{document}

\maketitle

\begin{abstract}
AI-driven content generation has made remarkable progress in recent years. However, neural networks and human designers operate in fundamentally different ways, making collaboration between them challenging. We address this gap for Scalable Vector Graphics (SVG) by equipping neural networks with tools commonly used by designers, such as axis alignment and explicit continuity control at command junctions. We introduce DesigNet, a hierarchical Transformer-VAE that operates directly on SVG sequences with a continuous command parameterization. Our main contributions are two differentiable modules: a continuity self-refinement module that predicts $C^0$, $G^1$, and $C^1$ continuity for each curve point and enforces it by modifying Bézier control points, and an alignment self-refinement module with snapping capabilities for horizontal or vertical lines.

DesigNet produces editable outlines and achieves competitive results against state-of-the-art methods, with notably higher accuracy in continuity and alignment. These properties ensure the outputs are easier to refine and integrate into professional design workflows. Source Code: \url{https://github.com/TomasGuija/DesigNet}.


\end{abstract}

\section{Introduction}
\label{sec:intro}
Typeface design is a fundamental case of vector graphics creation. Typefaces are ubiquitous in posters, books, user interfaces, and logos, with typography playing a decisive role in the visual identity of text. The professional font market is a multi-billion-dollar creative industry where subtle geometric choices often separate successful families from the rest.

In recent years, we have witnessed a rapid transformation in image generation, where convolutional and diffusion models are now capable of generating high-quality images with significant control. However, generating vector graphics entails fundamentally different challenges. The path-based representation makes spatial reasoning about style and composition harder, although the information is more compact than in images. Professional-level design requires consistency across key attributes, including weight, slant, width, and optical size. Technical constraints also play a vital role: expert designers minimize control points and place them at extremal positions to optimize rasterization and hinting~\cite{ahrens2012closer,hanover2020peter}. While prior automatic font-design methods have made significant progress~\cite{svg-vae-lopes2019, diffvg-Li:2020:DVG, DeepSVG-carlier2020, deepvecfont2_wang2023} through innovations in model architectures and learning objectives, they often drift from the intended style and produce vector outlines that designers find difficult to refine further.

Our main idea is to equip the model with the same high-level controls that designers use in practice. Designers typically specify whether a segment is a straight line or a cubic Bézier curve. They set continuity at junctions, from $C^0$ (only geometric continuity) through $G^1$ (collinear tangents) to $C^1$ (collinear tangents with equal magnitude)~\cite{prautzsch2002bezier}, illustrated in Fig. \ref{fig:Cont-types}.

We introduce differentiable continuity and alignment self-refinement modules that expose geometric design decisions to the network via deterministic geometric operators.  In addition to predicting standard drawing commands and their arguments, the model predicts (i) a continuity label for each command and (ii) an axis-alignment label for each line segment. The corresponding operators then adjust control points to enforce the predicted continuity level and snap line segments to the predicted horizontal/vertical axes.

\begin{figure}
    \centering

    \includegraphics[width=\columnwidth]{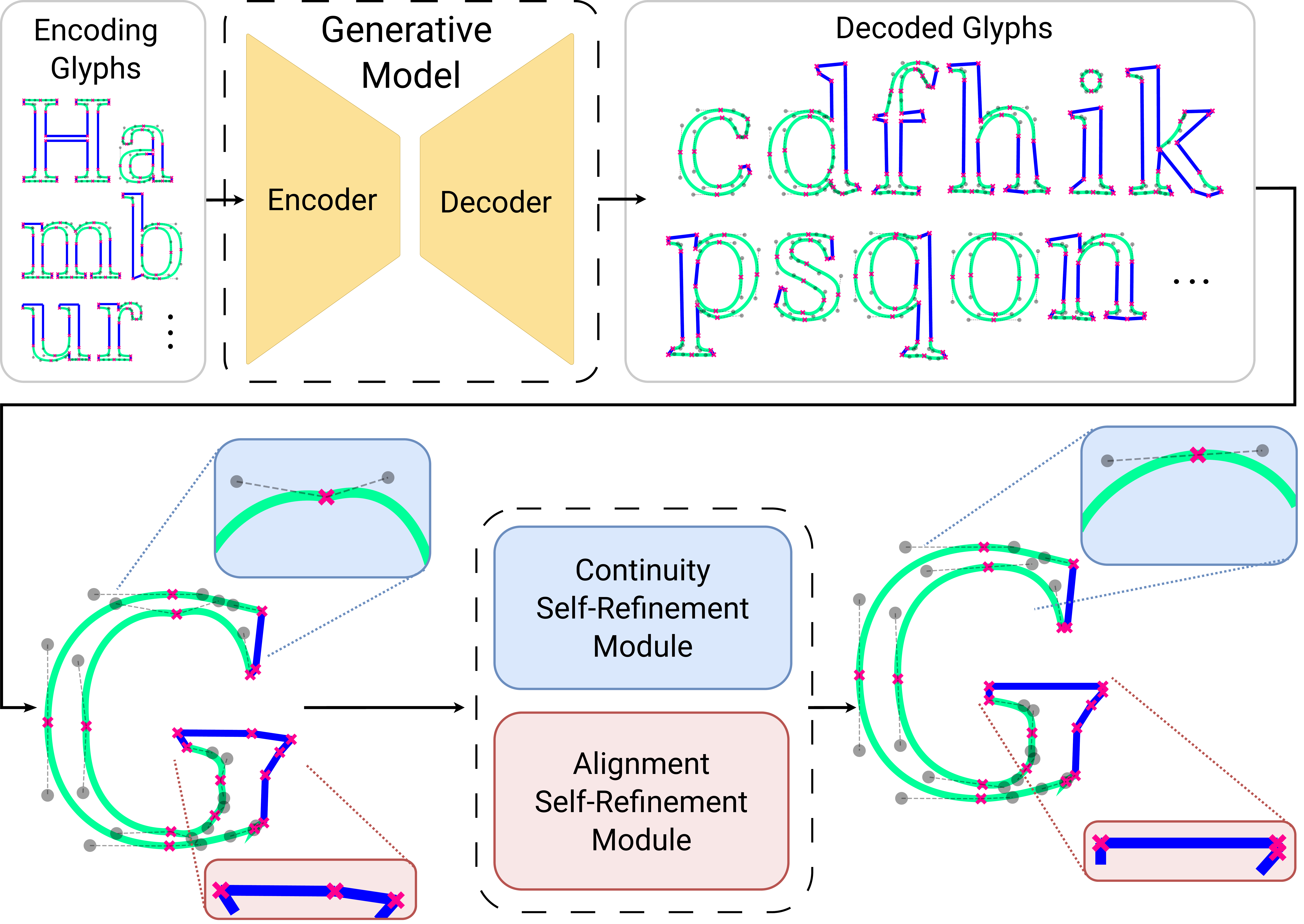}
    \caption{\textbf{Overview of DesigNet}. A subset of characters ("H", "a", "m", "b", "u", "r", ...) from a font is encoded to extract style features. The decoder then generates the remaining glyphs by combining the learned style with the embedding of the target letters. Finally, our self-refinement modules adjust control points and endpoints to enhance continuity and axis alignment, yielding cleaner SVG outputs.}
    \label{fig:teaser}
\end{figure}

In the proposed modules, discrete decisions are optimized end-to-end using straight-through estimators, enabling gradients to flow through the refinement process while applying hard geometric constraints in the forward pass. During training, both modules are supervised with ground-truth continuity and alignment labels and can be integrated into any SVG generator that predicts drawing commands. The resulting vector outlines preserve the target style while remaining structurally clean and easy to edit (Fig.~\ref{fig:teaser}).

We extend prior work on neural vector graphics~\cite{DeepSVG-carlier2020,deepvecfont2_wang2023} by replacing discrete drawing tokens with a continuous parameterization that mirrors a designer’s canvas. Discrete tokens lack smoothness guarantees, whereas a continuous parameterization supports smooth conditioning and decoding.

In the experimental section, we conduct an extensive evaluation across multiple datasets, including our internal Latin typeface dataset, a Chinese Fonts dataset~\cite{deepvecfont2_wang2023}, and an Icons dataset~\cite{DeepSVG-carlier2020}. Quantitative results demonstrate competitive performance with respect to state-of-the-art methods in terms of Intersection-over-Union, reconstruction error, and image-level $\ell_1$ distance. Qualitative evaluations further show that our model produces clean strokes, smooth curve transitions, and consistent stylistic attributes across the full alphabet. In addition, we show that the learned latent space enables smooth interpolations between entire fonts. Moreover, the learned latent space enables smooth interpolations between entire fonts. Most importantly, our proposed self-refinement modules significantly improve continuity and alignment accuracy, while remaining fully compatible with continuous coordinate representations.

Our main contributions are as follows:
\begin{itemize}

    \item We introduce novel continuity and alignment self-refinement modules that can be integrated into any SVG generator network, providing explicit supervision and enhanced control over continuity and alignment predictions.
    \item We replace the discrete input/output arguments with a continuous space and positional encodings, avoiding quantization loss and guaranteeing smoothness.
    \item We design a partitioned latent space that preserves fine-grained path-level detail while maintaining a coherent global font style.
    \item We extend the VAE to font-level generation, reconstructing an entire alphabet from a few reference glyphs, and evaluate it quantitatively and qualitatively.
\end{itemize}

\section{Related Work}
\label{sec:related-work}

A straightforward approach to font modeling is to leverage powerful generative models to synthesize raster images and subsequently trace them into vector graphics. However, despite the availability of smart tracing algorithms~\cite{selinger2003potrace, dominici2020polyfit, reddy2021im2vec, rodriguez2025starvector}, achieving the level of quality required for professional design requires direct supervision in the vector graphics domain.

Fortunately, several principles from image generation transfer to SVG modeling. Variational Autoencoders (VAEs)~\cite{kingma2013autoencodingvariationalbayes, kingma2019introduction} are easy to train but often suffer from blurry reconstructions. Generative adversarial networks (GANs)~\cite{goodfellow2014gans, creswell2018generative} improve visual fidelity at the cost of increased training instability and reduced controllability. Diffusion models~\cite{ho2020ddpm, songscore} also offer stable training but require expensive inference and are difficult to control precisely. We adopt a VAE formulation to enable smooth latent interpolations, font retrieval, and strict control over the generation process.

Early efforts operating directly in the SVG domain demonstrated the feasibility of learning generative models.

SVG-VAE~\cite{svg-vae-lopes2019} was the first method capable of generating new unseen glyphs by capturing font style from a small set of examples. It employs a class-conditioned, convolutional VAE to extract the style representation of the font and uses an LSTM decoder to generate the drawing commands of target glyphs.

DeepSVG~\cite{DeepSVG-carlier2020} was the first work to use a Transformer~\cite{vaswani2017attention} encoder-decoder architecture operating directly in SVG space. Their simple method combines two different encoders-decoders: one that operates independently for each path and a second that aggregates information across the entire glyph.

Building upon these ideas, DeepVecFont~\cite{deepvecfont_wang2021} and its improved variant DeepVecFont-v2~\cite{deepvecfont2_wang2023} specifically target the problem of font generation by adopting a dual-path architecture that processes raster images and vector representations in parallel, integrating features across both domains to enable glyph reconstruction.
DualVector~\cite{liu2023dualvectorunsupervisedvectorfont} also adopts a dual-path design, where raster and vector branches are modeled independently, with the raster prediction subsequently guiding a refinement stage applied to the generated vector glyphs.

More recently, diffusion-based approaches such as VecFusion~\cite{vecfusion-thamizharasan2024} have further advanced the state of the art, while interactive frameworks like FontCraft~\cite{fontcraft-Tatsukawa_2025} propose multimodal solutions for font design.
These works show that treating fonts as structured vector sequences can yield scalable and stylistically coherent generation, but challenges remain in preserving geometric consistency and fine-grained details.

SVGFormer~\cite{cao2023svgformer} is a transformer encoder-decoder that directly ingests the continuous SVG commands and combines them with positional information from the sequence and semantic labels generated from the Medial Axis Transform, alongside a redesigned attention mechanism tailored to vector graphics.

Beyond fonts, several works have explored the generation of more complex SVG images. Differentiable rendering approaches such as DiffVG~\cite{diffvg-Li:2020:DVG} enable gradient-based optimization over vector primitives, paving the way for learning in the SVG space. More recent contributions, including NIVeL (Neural Implicit Vector Layers) \cite{nivel-thamizharasan2024nivelneuralimplicitvector} and Neural Path Representation methods \cite{zhang2024texttovectorgenerationneuralpath}, address text-to-vector generation with higher fidelity. Diffusion-based approaches like VectorFusion \cite{vectorfusion-jain2022vectorfusiontexttosvgabstractingpixelbased}, SVGDreamer \cite{svgdreamer-10909425}, and related works demonstrate the potential of large-scale generative models to produce diverse SVG illustrations guided by textual prompts. These techniques highlight the generalizability of vector generation beyond the font domain.

Other interesting works based on Transformers are those that operate in point clouds~\cite{zhao2021pointtransformer}, and even spline-based geometries~\cite{splinebasedtransformers-Chandran_2024}.

Prior work has made significant progress in SVG generation for both fonts and general vector graphics. Yet, existing models face two persistent challenges: (i) capturing local geometric regularities such as continuity and alignment, which are essential for professional design, and (ii) maintaining global stylistic consistency across entire alphabets. Moreover, most approaches rely on discretized coordinate representations, which inevitably lose geometric precision. Our work addresses these limitations by combining hierarchical VAE modeling with Transformer-based encoding and decoding, together with explicit geometric supervision, enabling both fine-grained local control and coherent global style.

\section{Method}
\label{sec:method}

This section describes the components of DesigNet, a model that generates scalable vector fonts that combine the stylistic coherence of a given typeface with the precise geometric regularities valued by designers. We build upon a hierarchical VAE framework equipped with Transformer encoders and decoders.

A key design choice is the command representation. Prior work discretized coordinates, which simplifies enforcing exact horizontal and vertical lines as well as point alignment, but reduces precision and can introduce artifacts. We instead adopt continuous coordinates to preserve geometric accuracy and enable smooth interpolation, while enforcing geometric constraints through \emph{Self-Refinement} modules that adjust control points after decoding to satisfy continuity and axis alignment.

\subsection{Representation of SVG Data}
\label{sec:representation-svg-data}

We operate directly on typographic fonts represented in Scalable Vector Graphics (SVG) format, where each glyph is defined by one or more \texttt{<path>} elements containing a sequence of drawing commands. These commands describe cursor movements and geometric primitives such as straight lines and cubic Bézier curves, together with their coordinate arguments, which fully specify the contour of a character.

Formally, a glyph $\mathbf{G}_i$ is modeled as a collection of $N_p$ contours,
\begin{equation}
\mathbf{G}_i = [\mathbf{P}_{i1}, \mathbf{P}_{i2}, \dots, \mathbf{P}_{iN_p}],
\end{equation}
where each contour path $\mathbf{P}_{ij}$ is a sequence of $N_c$ commands,
\begin{equation}
\mathbf{P}_{ij} = [C_{ij1}, C_{ij2}, \dots, C_{ijN_c}].
\end{equation}

Each command $C_{ijk}$ is represented as a tuple $(z_{ijk}, \mathbf{A}_{ijk})$, where $\mathbf{A}_{ijk}$ are its coordinate arguments and $z_{ijk} \in \mathcal{Z} = \{ \mathtt{MoveTo}, \mathtt{LineFromTo}, \mathtt{CurveFromTo}, \mathtt{EOS} \}$ denotes the command type. These labels correspond to the standard primitives used in professional font formats and SVG path descriptions, with $\mathtt{EOS}$ marking the end-of-sequence.

To unify lines and curves under a common representation, we adopt the 4-points parameterization of~\cite{deepvecfont2_wang2023},
\begin{equation}
\mathbf{A}_{ijk} = (\mathbf{p}^1_{ijk}, \mathbf{p}^2_{ijk}, \mathbf{p}^3_{ijk}, \mathbf{p}^4_{ijk}), \quad \mathbf{p} \in \mathbb{R}^2.
\end{equation}
For cubic Bézier curves represented with the command $\mathtt{CurveFromTo}$, $\mathbf{p}^1_{ijk}$ and $\mathbf{p}^4_{ijk}$ denote the start and end points, while $\mathbf{p}^2_{ijk}, \mathbf{p}^3_{ijk}$ are the control points. For $\mathtt{LineFromTo}$ and $\mathtt{MoveTo}$ commands, only $(\mathbf{p}^1_{ijk}, \mathbf{p}^4_{ijk})$ are used.
Note also that this 4-points parameterization is redundant, because ideally, the network should produce $\mathbf{p}^4_{ijk-1} = \mathbf{p}^1_{ijk}$. This redundancy is introduced to make each command self-contained, which simplifies the learning process. To ensure the consistency of $\mathbf{p}^4_{ijk-1}$ and $\mathbf{p}^1_{ijk}$ predictions we use a specific loss term detailed in Section \ref{sec:Loss}.

Finally, for efficient training, we pad each glyph to a fixed maximum number of contours $N_p^{\text{max}}$ and each contour to a maximum number of commands $N_c^{\text{max}}$. Empty slots are padded with $\mathtt{EOS}$ commands and dummy coordinates, yielding a fixed tensor representation of shape $(N_p^{\text{max}}, N_c^{\text{max}}, 4, 2)$ for every glyph, which simplifies batching.

\subsection{Continuous SVG Embeddings}
\label{sec:svg-embedding}

Each drawing command is projected into a shared $d_E$-dimensional embedding space before being processed by the network. Specifically, a command $C_{ijk}$ is mapped to

\begin{equation}
\mathbf{e}_{ijk} \;=\;
\underbrace{\mathbf{E}_{\mathrm{cmd}}(z_{ijk})}_{\text{command type}}
+ \underbrace{f_{\mathrm{arg}}\!\left(\mathbf{A}_{ijk}\odot \mathbf{M}_{ijk}\right)}_{\text{argument embedding}}
+ \underbrace{\mathrm{PE}(k)}_{\text{positional encoding}},
\end{equation}

where $\mathbf{E}_{\mathrm{cmd}}$ is a learnable lookup table that assigns each command type a vector in $\mathbb{R}^{d_E}$, $f_{\mathrm{arg}}$ embeds the continuous arguments, and $\mathrm{PE}(k)$ is a sinusoidal positional encoding of the command index $k$.

The argument embedding $ f_{\mathrm{arg}}$ is implemented as a linear layer where $\mathbf{A}_{ijk}\in \mathbb{R}^{4\times 2}$ is the 4-point representation of command arguments and $\mathbf{M}_{ijk}\in \{0,1\}^{4\times 2}$ is a binary mask that removes padded values for $\mathtt{MoveTo}$ and $\mathtt{LineFromTo}$ commands. Unlike prior works that discretize coordinates~\cite{DeepSVG-carlier2020, deepvecfont2_wang2023}, we preserve continuous arguments, avoiding quantization artifacts and enabling finer geometric precision.

\subsection{Hierarchical Transformer-VAE}
\label{sec:hierarchical-transformer-vae}

DesigNet architecture is inspired by DeepSVG~\cite{DeepSVG-carlier2020}, which introduced a Transformer-based encoder–decoder for scalable vector graphics. DeepSVG demonstrated the feasibility of modeling SVG commands directly by using a two-level hierarchy of encoders and decoders. The first level processes each path individually across the sequence dimension, and the second aggregates information from different paths to produce a single latent code per glyph.

While effective for generic SVG generation, its design compresses the entire glyph into a single latent code and relies on discretized coordinates, which limits geometric precision and reconstruction fidelity in the context of font modeling. In contrast, our approach introduces a partitioned latent space with both global and path-level latents, continuous coordinate embeddings, and explicit geometric supervision. These design choices enable fine-grained local control while preserving global stylistic coherence.

\noindent\textbf{Encoder.}
The encoder maps an input glyph to a latent distribution using a two-stage hierarchical design. Since DeepSVG considers only one glyph at a time, in this subsection we will drop the glyph index $i$ for simplicity.

\begin{enumerate}
    \item \textbf{Path-level encoding.}
    Each path $\mathbf{P}_j$ is processed independently by a Transformer encoder $E^{(1)}$, producing contextualized command embeddings $\{\mathbf{\tilde{e}}_{jk}\}_{k=1}^{N_c}$. We summarize each path by average pooling:
    \begin{equation}
        \mathbf{u}_j = \frac{1}{N_c} \sum_{k=1}^{N_c} \mathbf{\tilde{e}}_{jk},
    \end{equation}
    yielding a path embedding $\mathbf{u}_j \in \mathbb{R}^{d_E}$ that captures local geometric structure.

    \item \textbf{Glyph-level encoding.}
    The set of path embeddings $\{\mathbf{u}_{j}\}_{j=1}^{N_p}$ is augmented with sinusoidal positional encodings and passed through a second Transformer encoder $E^{(2)}$, which models dependencies across paths. A visibility-aware average pooling operation aggregates the outputs into a glyph-level embedding $\mathbf{g} \in \mathbb{R}^{d_E}$, from which the encoder predicts the parameters of a Gaussian distribution $\mathcal{N}\left(\hat{\boldsymbol{\mu}}, \hat{\boldsymbol{\sigma}}\right)$.
\end{enumerate}

\noindent\textbf{Partitioned latent space.}
Inspired by NVAE~\cite{nvae-vahdat2021nvaedeephierarchicalvariational}, and to avoid compressing all information into a single vector, we extend the architecture with a partitioned latent space. In addition to a global latent $\mathbf{z} \in \mathbb{R}^{d_z}$, we introduce path-level latents $\{\mathbf{z}_j \in \mathbb{R}^{d_z}\}_{j=1}^{N_p}$:

\begin{equation}
\begin{aligned}
    \mathbf{z}   &= \hat{\boldsymbol{\mu}}   + \hat{\boldsymbol{\sigma}}   \odot \boldsymbol{\epsilon},   &\quad \boldsymbol{\epsilon}   &\sim \mathcal{N}(\mathbf{0}, \mathbf{I}), \\
    \mathbf{z}_j &= \hat{\boldsymbol{\mu}}_j + \hat{\boldsymbol{\sigma}}_j \odot \boldsymbol{\epsilon}_j, &\quad \boldsymbol{\epsilon}_j &\sim \mathcal{N}(\mathbf{0}, \mathbf{I}).
\end{aligned}
\end{equation}

The parameters $(\hat{\boldsymbol{\mu}}_j, \hat{\boldsymbol{\sigma}}_j)$ are predicted from each path embedding $\mathbf{u}_j$. This design preserves fine-grained path-level information while ensuring probabilistic regularization at both local and global scales.

\noindent\textbf{Decoder.}
The decoder mirrors the hierarchy of the encoder.

\begin{enumerate}
    \item \textbf{Path-level decoding.}
    The path-level latents $\mathbf{z}_j$ are processed by a Transformer decoder $D^{(2)}$ with cross-attention to the global latent $\mathbf{z}$, conditioning local geometry on global style and producing refined path embeddings $\{\mathbf{\hat{u}}_j \in \mathbb{R}^{d_E}\}$. An MLP predicts auxiliary attributes such as a scalar visibility logit $\hat{v}_j \in \mathbb{R}$.

    \item \textbf{Command-level decoding.}
    For each path, sinusoidal positional encodings serve as a fixed query template that defines the command order. A Transformer decoder $D^{(1)}$ attends to the refined path embedding $\mathbf{\hat{u}}_j$, producing contextualized command embeddings $\{\hat{\mathbf{e}}_{jk}\}$. A final MLP outputs the command type, continuous arguments, continuity, and alignment logits, which we will explain in the following sections.
\end{enumerate}

This hierarchical partitioned VAE balances global coherence with local precision: global latents capture overall style, while path-level latents preserve detailed geometry. The result is a well-regularized yet expressive generative process that produces geometrically precise and stylistically coherent SVG glyphs.

\subsection{Continuity Self-Refinement Module}

We explicitly model the geometric continuity at the junctions between consecutive segments within a glyph contour. In our representation, each segment is either a line or a cubic Bézier curve. Consecutive line segments are restricted to $C^0$ continuity by construction, while cubic Bézier curves can exhibit higher-order continuities depending on their control points. We distinguish three continuity levels (see Fig. \ref{fig:Cont-types} for a graphical illustration):

\begin{figure}[t]
    \centering
    \includegraphics[width=\columnwidth]{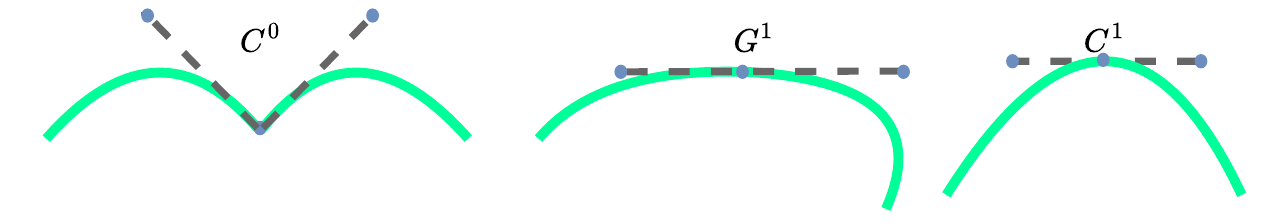}
    \vspace{-0.6cm}
    \caption{\textbf{Illustration of different continuity types}. $C^0$: only geometric continuity, $G^1$: collinear tangents, and $C^1$: collinear tangents with equal magnitude.}
    \vspace{-0.4cm}
    \label{fig:Cont-types}
\end{figure}

\begin{itemize}
    \item \textbf{$C^0$ continuity.}
    Segments share a common endpoint. By construction, all contiguous segments satisfy $C^0$ continuity.

    \item \textbf{$G^1$ continuity.}
    Tangent directions at the shared endpoint are collinear. For a line, the tangent is the vector from start to end; for a cubic Bézier, it is the vector from the endpoint to its nearest control point. Let $\mathbf{t}^-$ and $\mathbf{t}^+$ denote the outgoing and incoming tangents. We consider the junction to have $G^1$ continuity if
    \begin{equation}
        \frac{\mathbf{t}^- \cdot \mathbf{t}^+}{\|\mathbf{t}^-\|\,\|\mathbf{t}^+\|} > 1 - \epsilon_a .
    \end{equation}

    \item \textbf{$C^1$ continuity.}
    In addition to $G^1$, a junction will have $C^1$ continuity if the length of their tangents is equal: $\left| \norm{\mathbf{t}^-} - \norm{\mathbf{t}^+} \right| < \epsilon_b$.
\end{itemize}

Here, $\epsilon_a$ and $\epsilon_b$ are small positive thresholds that account for numerical imprecision and approximate continuity in learned vector representations.

For the last endpoint $\mathbf{p}^4_{jk}$ of the command $C_{jk}$, the model predicts a distribution over continuity labels $\hat{y}_{jk} \in \{\text{$C^0$}, \text{$G^1$}, \text{$C^1$}\}$, with supervision from ground-truth labels $y_{jk}$ computed directly from SVG geometry.

We exploit these predictions by applying a deterministic geometric refinement step that adjusts Bézier control points so that junctions satisfy the predicted level of continuity.
During training, this refinement is implemented as a differentiable module using a straight-through estimator~\cite{DBLP:journals/corr/BengioLC13,yin2019understanding}: the forward pass applies the hard predicted label via an argmax operation, while gradients are propagated through a softmax relaxation. At inference time, the same refinement is applied using hard decisions only.

For line–curve junctions, we modify only the control point of the Bézier curve adjacent to the shared endpoint, moving it so that the curve tangent at the junction aligns with the line direction.

For curve-curve junctions, let $\widehat{\mathbf{t}}^-$ and $\widehat{\mathbf{t}}^+$ be the normalized tangents. In the case of $G^1$, we correct the control points of the two curves adjacent to the junction—specifically, the control point preceding the endpoint of the first curve and the control point following the start point of the second curve—such that the updated tangents satisfy:

\begin{equation}
\mathbf{t}^{-} = -\norm{\mathbf{t}^-} \mathbf{d}; \quad
\mathbf{t}^{+} = \norm{\mathbf{t}^+} \mathbf{d}; \quad
\mathbf{d} =
\frac{\widehat{\mathbf{t}}^- + \widehat{\mathbf{t}}^+}
{\norm{\widehat{\mathbf{t}}^- + \widehat{\mathbf{t}}^+}}
\end{equation}

To enforce $C^1$, we not only impose a common direction but also a common norm of the tangents:
\begin{equation}
\mathbf{t}^{-} = -s \mathbf{d}; \quad
\mathbf{t}^{+} = s \mathbf{d}; \quad
s = \frac{\norm{\mathbf{t}^+} + \norm{\mathbf{t}^-}}{2}.
\end{equation}

This symmetric adjustment modifies both adjacent control points equally, aligning tangent directions for $G^1$ continuity and both directions and magnitudes for $C^1$ continuity. By coupling continuity prediction with geometric refinement, the module enhances the smoothness and stylistic coherence of generated glyphs without requiring additional parameters, ensuring compatibility with professional font-editing tools.

\subsection{Alignment Self-Refinement Module}

In addition to continuity, a common operation in font design is to set the same $x$ or $y$ coordinate of path points in order to achieve vertical or horizontal alignment. We model this operation by predicting an axis-alignment class for every line drawing command. For a line segment with start $\mathbf{a}=\left[x_s,y_s\right]^\top$ and end $\mathbf{b}=\left[x_e,y_e\right]^\top$, the model outputs alignment logits defining a distribution over $\hat{\alpha} \in \{\texttt{H}, \texttt{V}, \varnothing\}$, corresponding to horizontal, vertical, or no alignment. Ground-truth alignment labels $\alpha$ are computed directly from the SVG geometry.

These predictions are exploited through a deterministic alignment refinement step that snaps line segments predicted as horizontal or vertical to the corresponding axis. Only line endpoints are modified; no additional parameters are introduced.
As in the continuity refinement module, this alignment refinement is implemented as a differentiable module using a straight-through estimator~\cite{DBLP:journals/corr/BengioLC13}.

At inference time, the same refinement is applied using hard decisions only. Let $\bar{x} = \tfrac{1}{2}(x_s + x_e)$ and $\bar{y} = \tfrac{1}{2}(y_s + y_e)$. The snapped endpoints $\left(\mathbf{a}',\mathbf{b}'\right)$ are defined as
\begin{equation}
\left(\mathbf{a}',\mathbf{b}'\right) =
\begin{cases}
\left((x_s,\bar{y}), (x_e,\bar{y})\right), & \text{if } \hat{\alpha} = \texttt{H}, \\[2pt]
\left((\bar{x},y_s), (\bar{x},y_e)\right), & \text{if } \hat{\alpha} = \texttt{V}, \\[2pt]
\left((x_s,y_s), (x_e,y_e)\right), & \text{if } \hat{\alpha} = \varnothing.
\end{cases}
\end{equation}

This snapping procedure reduces orientation noise in thin strokes and improves the crispness of horizontal and vertical structures (e.g., crossbars and stems), thereby improving editability and compatibility with professional font design tools.

\begin{figure}[t]
    \centering
    \includegraphics[width=0.95\columnwidth]{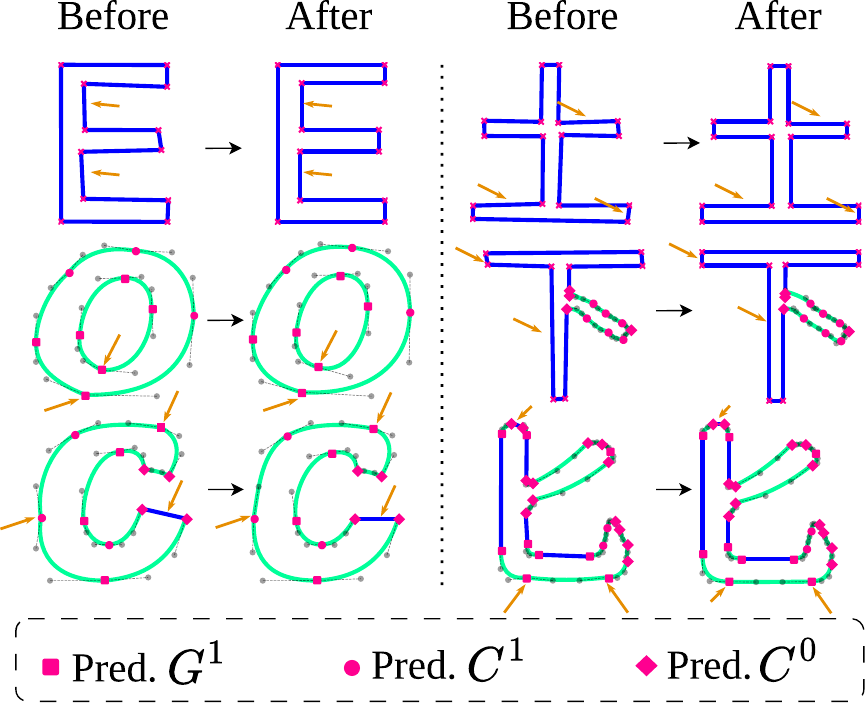}
    \caption{\textbf{Examples of Continuity and Alignment Modules}. We show the predicted glyphs before and after the self-refinement modules. $\mathtt{LineFromTo}$ commands are shown in blue and $\mathtt{CurveFromTo}$ commands in green. The predicted continuity is represented by the pink squares, circles, and diamonds. We highlight with orange arrows the junctions corrected by the Self-refinement modules based on the pink predictions.}
    \label{fig:self_refinement_samples}

\end{figure}

\subsection{DesigNet: A Font Generative Model}

We build a font generator on top of our hierarchical VAE architecture. The model takes a fixed set of \emph{encoding glyphs} (reference letters) and generates \emph{decoding glyphs} (held-out letters) in the same style.

Given $N_{\mathrm{enc}}$ reference glyphs $\{\mathbf{G}_i\}_{i=1}^{N_{\mathrm{enc}}}$, their command sequences are first processed by the VAE encoder just as described in Sec. \ref{sec:hierarchical-transformer-vae}. At the \emph{path level}, $E^{(1)}$ produces path embeddings $\mathbf{u}_{ij}$ for each individual path. To aggregate stylistic information across reference glyphs, we introduce $N_p$ learnable query vectors, one per path slot. A Transformer decoder attends from these queries to the set of path embeddings, while masking non-visible paths. The result is a collection of $N_p$ style-aware path representations, each encoding how the font style manifests in a specific path slot across glyphs. These vectors are then used to estimate the parameters of Gaussian distributions, from which path-level latent variables are sampled via the reparameterization trick.

At the \emph{glyph level}, $E^{(2)}$ aggregates the path embeddings of each reference glyph into a corresponding glyph embedding. A Transformer with a \texttt{[CLS]} token processes these embeddings, producing a single glyph-level style representation. From this representation we derive the parameters of the global latent distribution $(\boldsymbol{\mu}, \boldsymbol{\sigma})$ and, using the reparameterization trick, sample a global style code $\mathbf{z} \in \mathbb{R}^{d_z}$.

In contrast to other methods, we adopt a partitioned latent space in which font style is encoded at both the glyph and contour levels. The global latent $\mathbf{z} \in \mathbb{R}^{d_z}$ captures overall stylistic properties such as weight, slant, or contrast, while the path-level latents $\{\mathbf{z}_{j} \in \mathbb{R}^{d_z}\}$ preserve localized geometric features that are crucial for reconstructing fine details of each contour. This separation alleviates the bottleneck of compressing an entire glyph into a single code, enabling global coherence and local geometric precision to be modeled simultaneously.

We steer generation to a target letter by conditioning the decoder on embeddings of the corresponding character identity. We concatenate these character embeddings into the decoder memory at both the path and command levels, allowing queries to attend to style latents and the target identity. Reference glyphs provide style, while target embeddings specify the character shape.

\subsection{Loss Function}
\label{sec:Loss}
Our training objective combines standard VAE reconstruction and regularization with several task-specific supervision terms.

First, we consider the reconstruction loss term $\mathcal{L}_{\text{rec}}$ as a composition of several loss components, following~\cite{deepvecfont2_wang2023}. In particular, we include a command type classification loss that enforces valid SVG command sequences, an argument regression loss that supervises the continuous parameters of each command while masking padded entries, and a visibility loss that determines whether each path is present in the glyph. In addition, we introduce an endpoint–start consistency loss that penalizes discrepancies between duplicated representations of junction coordinates, ensuring geometric coherence at command boundaries. Finally, an auxiliary rendering loss encourages alignment between predicted and ground-truth segments by comparing sampled points along each segment. Together, these terms enforce accurate geometry, structural validity, and visual fidelity at both the command and path levels.

Moreover, we incorporate KL regularization~\cite{kingma2013autoencodingvariationalbayes} to align the approximate posterior distributions with an isotropic Gaussian prior. In our hierarchical formulation, this regularization is applied at both the global glyph level and the path level, encouraging a compact and well-structured latent space.

Beyond these components, we incorporate our two main contributions: continuity and alignment supervision.
At the $i$-th joint between two commands, the model predicts a probability distribution $\hat{p}_i(c)$ over continuity classes $c \in \{C^0, G^1, C^1\}$, with ground-truth label $y_i$. We supervise this prediction using a cross-entropy loss. To account for the varying severities of continuity errors, we adopt a cost-sensitive formulation~\cite{krishnapuram2011cost} with a weight matrix $\mathbf{W} \in \mathbb{R}^{3\times 3}$ encoding misclassification costs, where confusions $C^0 \leftrightarrow C^1$ are penalized more heavily than confusions between $C^0 \leftrightarrow G^1$ and $G^1 \leftrightarrow C^1$:

\begin{equation}
\mathcal{L}_{\text{cont}}
= - \frac{1}{N_{\text{joints}}}
\sum_{i=1}^{N_{\text{joints}}}
\sum_{c \in \{C^0,G^1,C^1\}}
\mathbf{W}_{y_i,c} \,\log \hat{p}_i(c),
\end{equation}

For the $k$-th line segment, we additionally predict an alignment probability $\hat{p}_k(\alpha)$ over three categories $\alpha \in \{\texttt{H}, \texttt{V}, \varnothing\}$, with ground-truth label $\alpha_k$. This prediction is trained with a cross-entropy loss:

\begin{equation}
\mathcal{L}_{\text{align}}
= - \frac{1}{N_{\text{lines}}}
\sum_{k=1}^{N_{\text{lines}}}
\sum_{\alpha \in \{\texttt{H}, \texttt{V}, \varnothing\}}
 \mathbbm{1}\!\left[\alpha_k = \alpha\right]\log \hat{p}_k(\alpha),
\end{equation}

where $\mathbbm{1}[\cdot]$ is the indicator function. This term encourages the model to respect typographic regularities such as horizontal baselines and vertical stems. To sum up, our final loss is
\begin{equation}
\mathcal{L}_{\text{total}} \;=\;
\mathcal{L}_{\text{rec}}
+ \lambda_{\text{KL}}\,\mathcal{L}_{\text{KL}}
+ \lambda_{\text{cont}}\,\mathcal{L}_{\text{cont}}
+ \lambda_{\text{align}}\,\mathcal{L}_{\text{align}}.
\end{equation}

The reconstruction component $\mathcal{L}_{\text{rec}}$ enforces accurate geometry and visibility. $\mathcal{L}_{\text{KL}}$ keeps the latent space compact and allows for smooth interpolation, while $\mathcal{L}_{\text{cont}}$ and $\mathcal{L}_{\text{align}}$ encourage structural coherence in terms of smoothness and axis alignment, as enforced by our Self-Refinement modules. Crucially, because the self-refinement modules are integrated as differentiable components during training via straight-through estimators, incorrect continuity or alignment predictions lead to suboptimal geometric refinements and, in turn, higher reconstruction error. This coupling ensures that errors in geometric decisions are directly penalized by the reconstruction loss, providing a strong learning signal for the model to estimate meaningful continuity and alignment distributions.

\section{Experiments}
\label{sec:experiments}

In this section, we quantitatively and qualitatively evaluate the proposed approach. We first describe the evaluation protocol on our proprietary dataset and conduct ablation studies to assess the impact of key design choices. We also evaluate the generalization capabilities of our model on an icon dataset, extending the analysis beyond the font domain. Finally, we evaluate both our Variational Autoencoder (VAE) and the full font generation model on two benchmark tasks: Latin font generation and Chinese font generation.
\subsection{Dataset and Implementation Details}

For our Latin font generation experiments, we curated a dataset through a combination of automated filtering and manual refinement. The final collection comprises 16{,}165 fonts grouped into 5{,}134 typographic families. To prevent data leakage, all fonts belonging to the same family are assigned exclusively to a single split. The dataset is divided into 14{,}485 fonts for training, 842 for validation, and 838 for testing.

To train with continuous arguments, all coordinates are normalized by the font’s Units Per EM and recentered around the origin $(0,0)$. This normalization highlights geometric symmetries that the model can exploit.

Our initial architecture uses 4 encoders and 4 decoders with 8 attention heads each, model and latent dimensionality of 256, and feed-forward layers of size 512. Glyphs are capped at four paths with up to 32 commands per path, for a maximum sequence length of 64. We train using AdamW~\cite{loshchilov2017adamw} with an initial learning rate of $10^{-4}$, reduced on plateau, and a batch size of 64 until convergence. As in~\cite{deepvecfont2_wang2023}, $\mathcal{L}_{\text{rec}}$ contains multiple terms accounting for commands, arguments, visibility, consistency and auxiliary points. We use the following weights: $\lambda_{\text{KL}}$ linearly increases from $0$ to $10$ during the first 10K steps, $\lambda_{\text{cont}}=1.0$, and $\lambda_{\text{align}}=1.0$.

\subsection{Ablation Study}
\label{sec:ablation-study}
We evaluate our contributions by using the DeepSVG~\cite{DeepSVG-carlier2020} architecture as a baseline. Since DeepSVG is a pure VAE, we focus on reconstruction quality, assessing the model’s ability to reproduce the input glyph while preserving its visual appearance. As reported in Table~\ref{tab:vae_comparison}, we measure reconstruction performance using: (1) Intersection over Union (IoU) between rasterized predictions and ground-truth glyphs; (2) the $\ell_1$ distance between the corresponding rasterized images; and (3) the Reconstruction Error (RE), computed as the Chamfer distance between point clouds sampled from the predicted and ground-truth SVGs.

In addition to these reconstruction metrics, we also evaluate (4) the continuity accuracy at junctions

\begin{equation}
    \text{Acc}_\text{cont} = \frac{1}{N_{\text{joints}}} \sum_{i=1}^{N_{\text{joints}}} \mathbbm{1}\!\left[\argmax_c\, \hat{p}_i(c) = y_i\right],
\end{equation}

and (5) line alignment accuracy

\begin{equation}
    \text{Acc}_\text{align} = \frac{1}{N_{\text{lines}}} \sum_{k=1}^{N_{\text{lines}}} \mathbbm{1}\!\left[\argmax_\alpha\, \hat{p}_k(\alpha) = \alpha_k\right].
\end{equation}

When evaluating the self-refinement modules, we apply a confidence-based rule: geometric refinement is performed only when the predicted continuity or alignment label is assigned a probability greater than 75\%.

\begin{figure}[ht]
    \centering
    \scriptsize
    \setlength{\tabcolsep}{4pt}
    \renewcommand{\arraystretch}{1.2}
    \begin{tabular}{@{}m{0.15\columnwidth} m{0.8\columnwidth}@{}}
        \centering GT & \includegraphics[width=0.80\columnwidth]{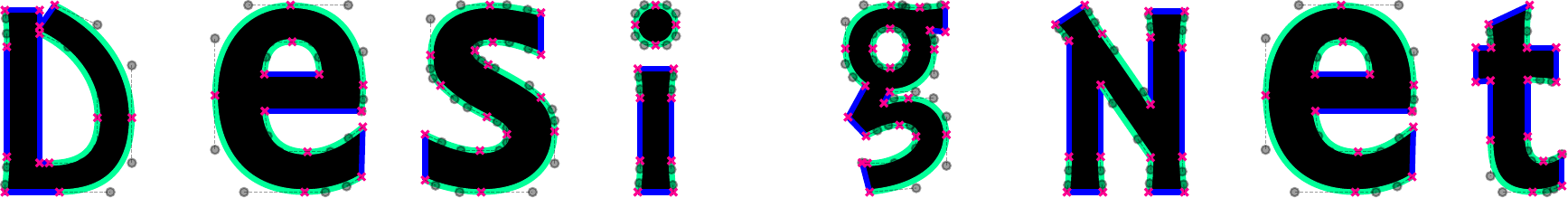} \\[4pt]
        \centering GT vs. Baseline \\(DeepSVG) & \includegraphics[width=0.8\columnwidth]{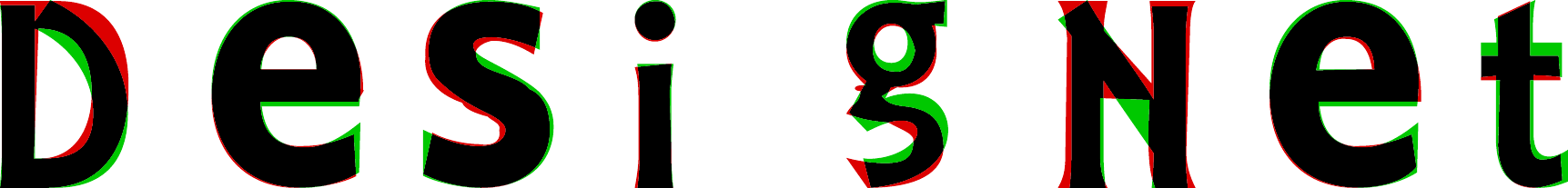} \\[4pt]
        \centering GT vs. ours (DesigNet) & \includegraphics[width=0.8\columnwidth]{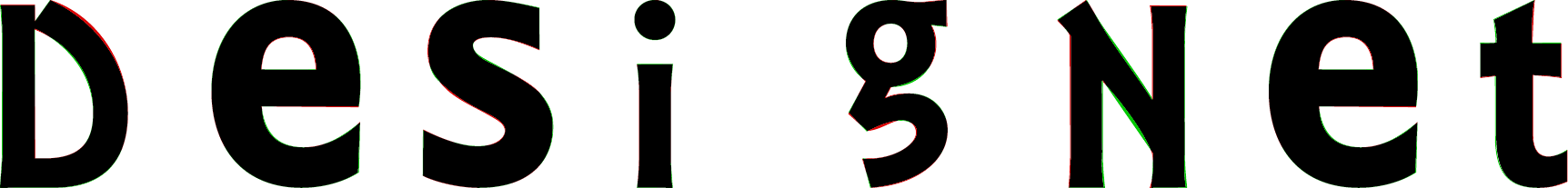} \\
    \end{tabular}
    \caption{
        \textbf{Qualitative comparison of reconstructed words using Latin fonts}. The first row presents the ground truth (GT) glyphs, including their joints and control points. The second and third rows compare the outputs of DeepSVG and DesigNet against the GT, where black indicates overlapping regions, green denotes GT regions not covered by the prediction, and red marks predicted regions that do not correspond to the GT.
    }
    \label{fig:qualitative_comparison}
\end{figure}

\begin{figure}[ht]
    \centering
    \scriptsize
    \setlength{\tabcolsep}{4pt}
    \renewcommand{\arraystretch}{1.2}
    \begin{tabular}{@{}m{0.15\columnwidth} m{0.8\columnwidth}@{}}
        \centering GT & \includegraphics[width=0.8\columnwidth]{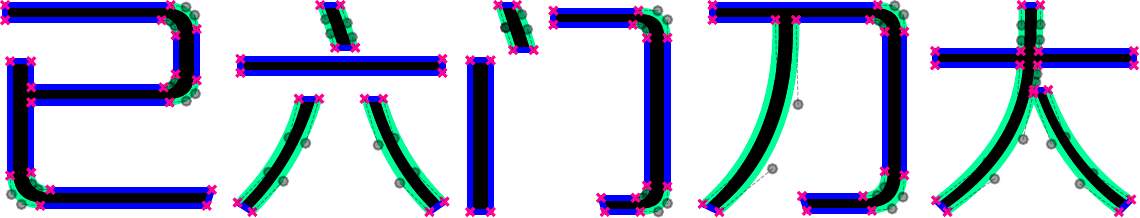
        } \\[4pt]
        \centering GT vs. Baseline \\(DeepSVG) & \includegraphics[width=0.8\columnwidth]{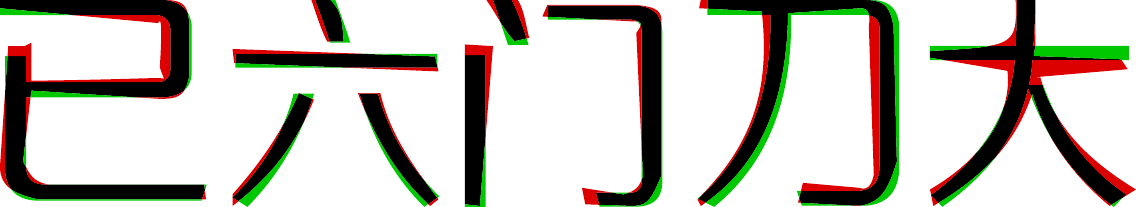} \\[4pt]
        \centering GT vs. ours (DesigNet) & \includegraphics[width=0.8\columnwidth]{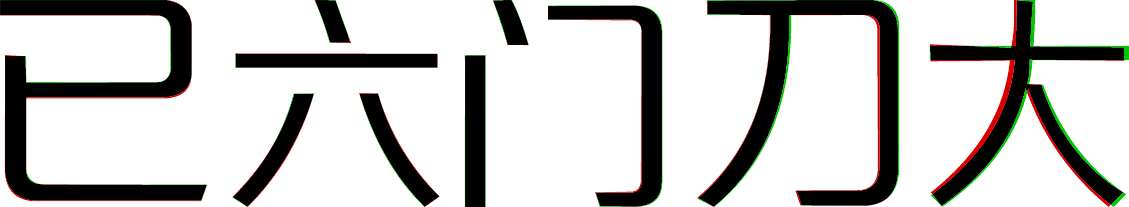} \\
    \end{tabular}
    \caption{
        \textbf{Qualitative comparison of reconstructed words using Chinese fonts}.
    }
    \label{fig:chinese_qualitative_comparison}
\end{figure}

Table~\ref{tab:vae_comparison} reports results on the test split of our curated dataset, where our method consistently outperforms the baseline across all metrics. Replacing discrete coordinates with continuous arguments yields a substantial improvement in reconstruction fidelity. Introducing a hierarchical latent space and a relaxed representation further enhances reconstruction accuracy. The addition of self-refinement modules produces a marked gain in continuity and alignment accuracy, underscoring their role in enforcing geometric regularity. Notably, without self-refinement, the Alignment Accuracy drops significantly when we introduce continuous arguments from 0.603 in DeepSVG to 0.368. This is natural, as discretization trivially enforces alignment when predictions fall within the same quantization bin. However, we prove that continuous representations achieve superior accuracy when combined with the proposed self-refinement modules.

\begin{table}[ht]
    \centering
    \resizebox{\columnwidth}{!}{
    \begin{tabular}{@{}p{0.32\columnwidth}ccccc@{}}
    \toprule
    \textbf{Model} & \textbf{IoU} $\uparrow$ & \textbf{L1} $\downarrow$ & \textbf{RE} $\downarrow$ & \textbf{Cont. Acc.} $\uparrow$ & \textbf{Align. Acc.} $\uparrow$\\
    \midrule
    DeepSVG & 0.789 & 0.069 & 8.782 &  0.567 & 0.603\\
    \midrule
    \makecell[l]{+ cont. args., \\ +sin. pos. enc.,\\ +centered glyphs}
               & 0.943 & 0.016 & 2.242 & 0.567 & 0.368\\
    \midrule
    \makecell[l]{+ hierarchical \\ latent space}
               & 0.963 & 0.010 & 1.477 & 0.651 & 0.380\\
    \midrule
    \makecell[l]{+ relaxed rep. \\ and aux. loss}
               & \textbf{0.970} & \textbf{0.009} &\textbf{1.137} & 0.686 & 0.376\\
    \midrule
    \makecell[l]{+ self-refinement \\ 75\% Conf. Trh.}
               & 0.969 & \textbf{0.009} & 1.138  & \textbf{0.886} & \textbf{0.969}\\
    \bottomrule
    \end{tabular}
    }
\caption{Ablation study of our VAE model on the test split of our proprietary dataset. Using continuous drawing parameters instead of the quantized representation in DeepSVG leads to a substantial improvement in reconstruction quality. Subsequent architectural enhancements further increase reconstruction fidelity, while the continuity and alignment self-refinement modules markedly improve continuity and alignment accuracy.}
\label{tab:vae_comparison}
\end{table}

For qualitative evaluation, we illustrate representative reconstructions for Latin and Chinese datasets in Figures~\ref{fig:qualitative_comparison} and~\ref{fig:chinese_qualitative_comparison}, respectively.

\subsection{Typeface Interpolation}
\begin{figure}
    \centering

    \setlength{\tabcolsep}{0pt}
    \renewcommand{\arraystretch}{0.0}

    \begin{tabular}{c}
        \includegraphics[width=0.48\textwidth]{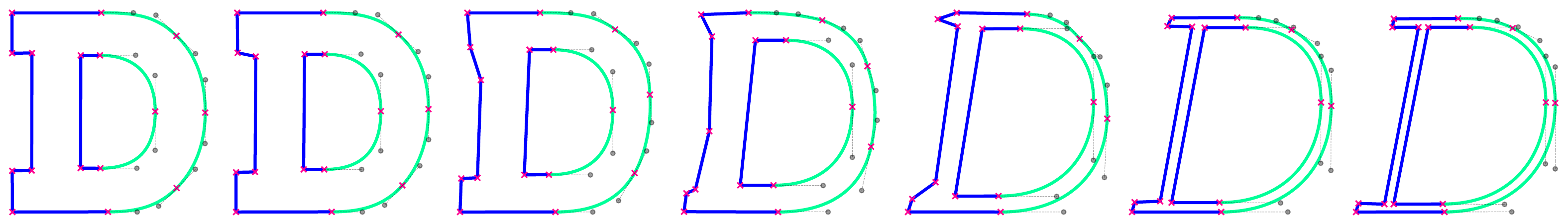} \\
        \vspace{0.2cm} \\
        \includegraphics[width=0.48\textwidth]{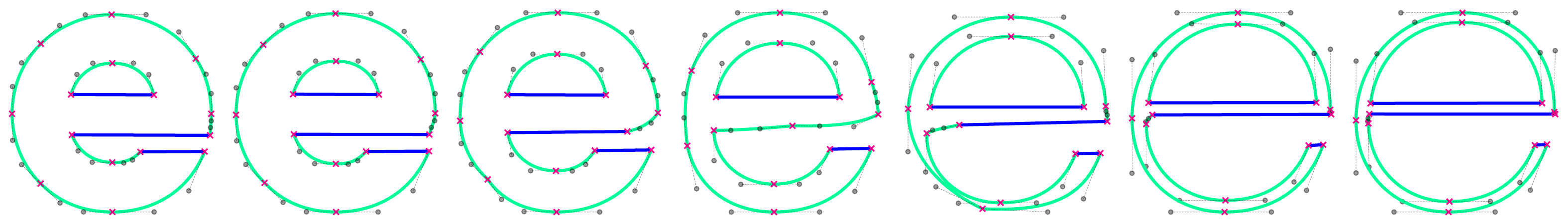} \\
        \vspace{0.2cm} \\
        \includegraphics[width=0.48\textwidth]{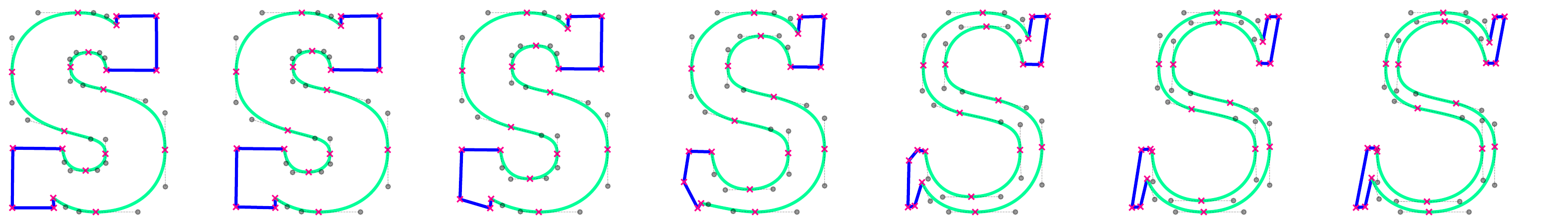} \\
        \vspace{0.2cm} \\
        \includegraphics[width=0.48\textwidth]{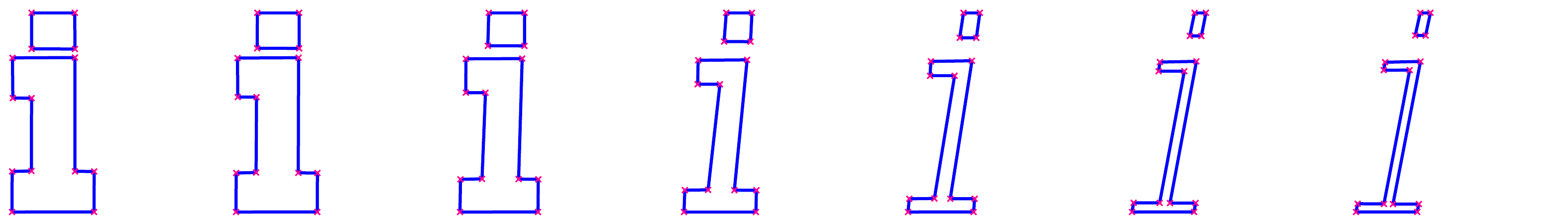} \\
        \vspace{0.2cm} \\
        \includegraphics[width=0.48\textwidth]{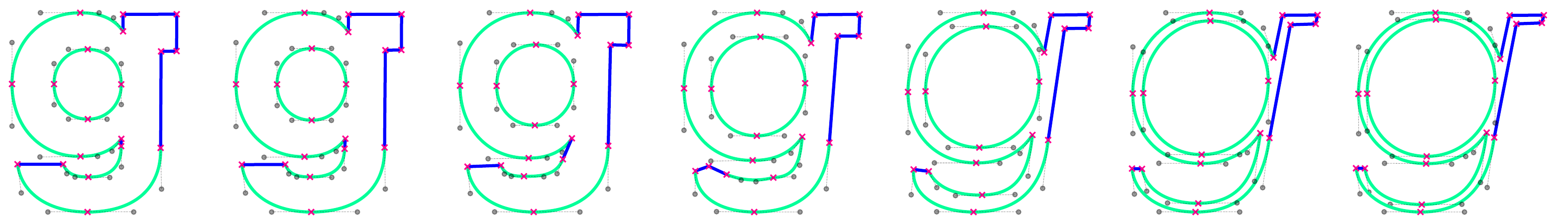} \\
        \vspace{0.2cm} \\
        \includegraphics[width=0.48\textwidth]{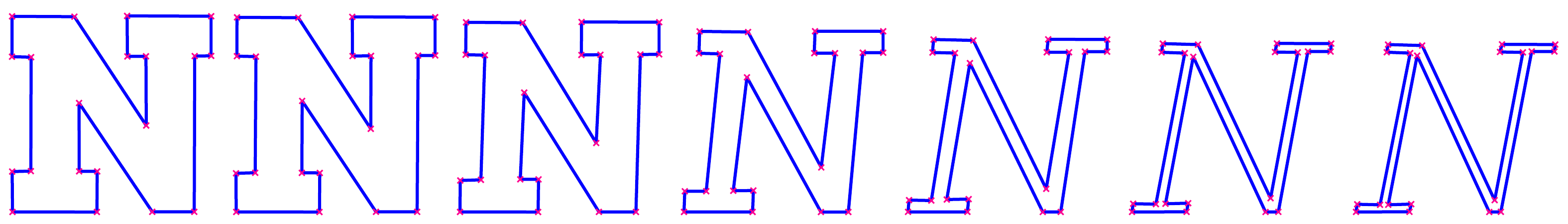} \\
    \end{tabular}

    \caption{\textbf{Font Interpolation}: Examples of latent space interpolation in DesigNet between two fonts with different weights and slants.}
    \label{fig:interpolations}
\end{figure}

Our model enables smooth interpolation between glyphs in the learned latent space. Figure~\ref{fig:interpolations} illustrates representative examples, showing gradual transitions between reconstructions of fonts with different weights and slants.

Given two glyphs with latent representations $\mathbf{z}_a$ and $\mathbf{z}_b$, we compute a linear interpolation
\begin{equation}
\mathbf{z}(\alpha) = (1 - \alpha) \, \mathbf{z}_a + \alpha \, \mathbf{z}_b, \quad \alpha \in [0, 1].
\end{equation}

Decoding $\mathbf{z}(\alpha)$ for varying values of $\alpha$ produces intermediate glyphs that smoothly transition between the two styles.

\subsection{Generalization with Icons}
To further assess the generalization capabilities of the proposed model, we qualitatively evaluate our approach on an icon dataset. This dataset is the same as the one used in DeepSVG \cite{DeepSVG-carlier2020} and is preprocessed following the same pipeline applied to our font dataset. In this setting, we do not predict fills or colors for the reconstructed icons; instead, we focus exclusively on reconstructing the vector paths defining their outlines, in line with our font generation setup.

\begin{figure}[ht]
    \centering
    \scriptsize
    \setlength{\tabcolsep}{3pt}
    \renewcommand{\arraystretch}{1.1}
    \setlength{\arrayrulewidth}{0.8pt}

    \begin{tabular}{@{}c c | c c@{}}
        \textbf{GT} & \textbf{Ours} & \textbf{GT} & \textbf{Ours} \\[2pt]

        \includegraphics[width=0.21\columnwidth]{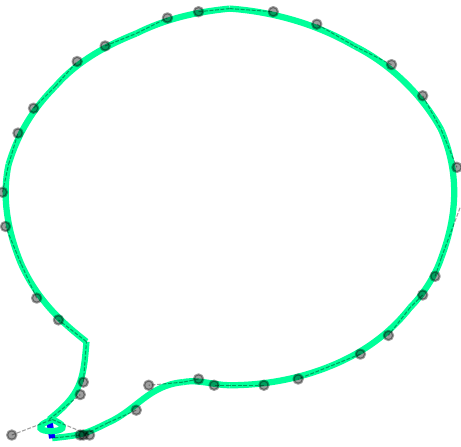} &
        \includegraphics[width=0.21\columnwidth]{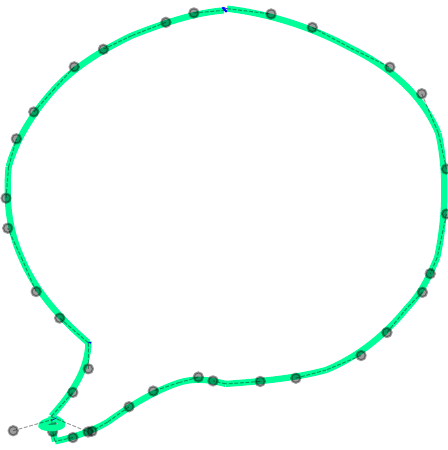} &
        \includegraphics[width=0.21\columnwidth]{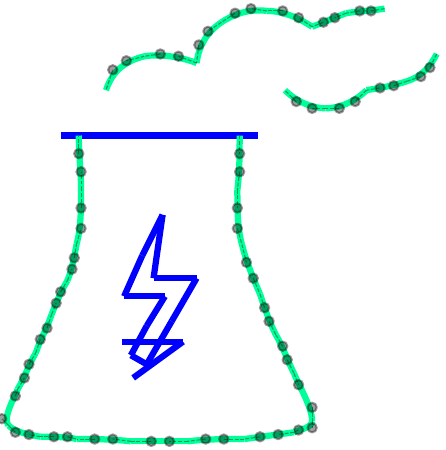} &
        \includegraphics[width=0.21\columnwidth]{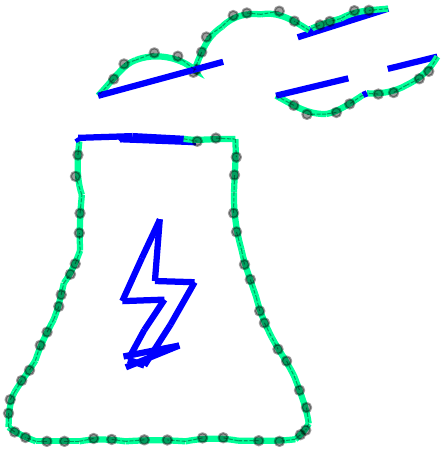} \\[4pt]

        \includegraphics[width=0.21\columnwidth]{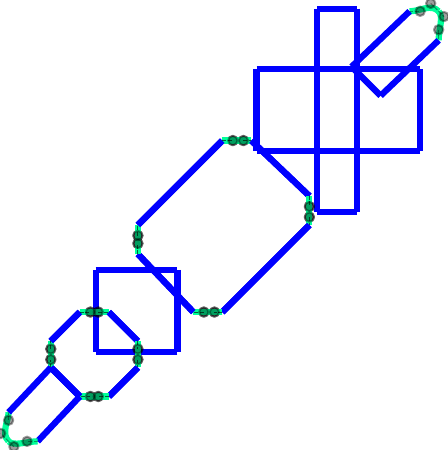} &
        \includegraphics[width=0.21\columnwidth]{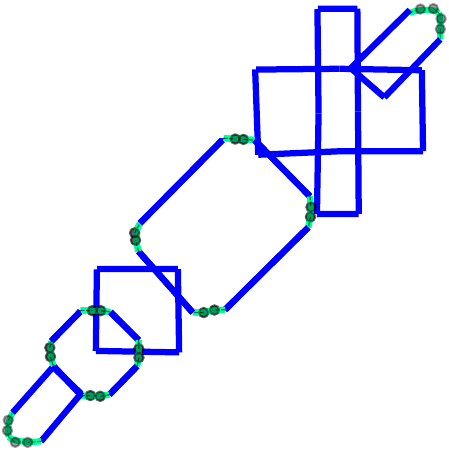} &
        \includegraphics[width=0.21\columnwidth]{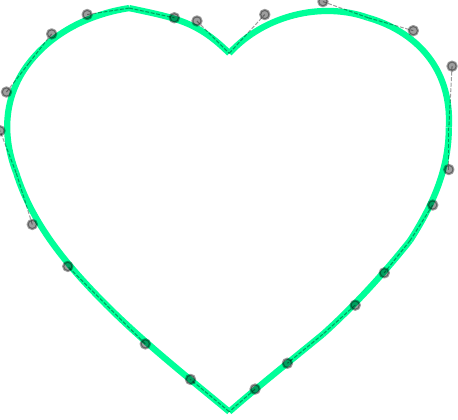} &
        \includegraphics[width=0.21\columnwidth]{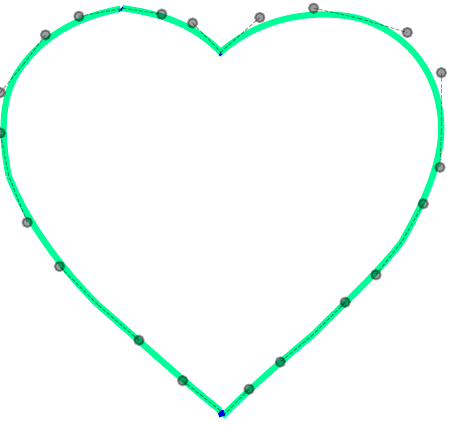} \\[4pt]

        \includegraphics[width=0.21\columnwidth]{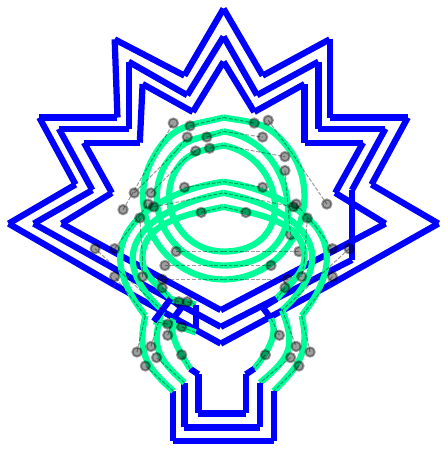} &
        \includegraphics[width=0.21\columnwidth]{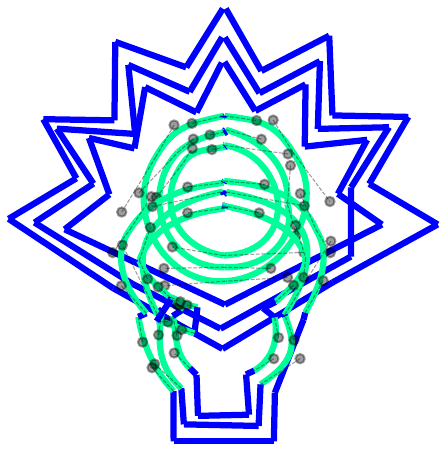} &
        \includegraphics[width=0.21\columnwidth]{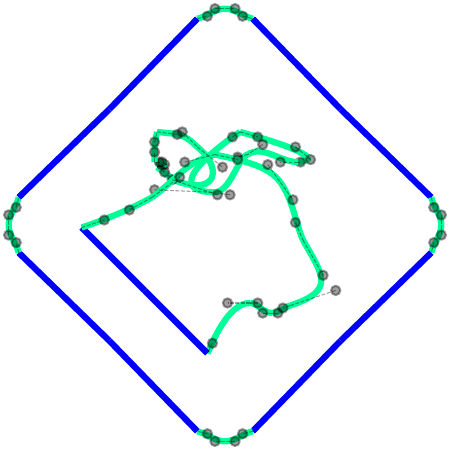} &
        \includegraphics[width=0.21\columnwidth]{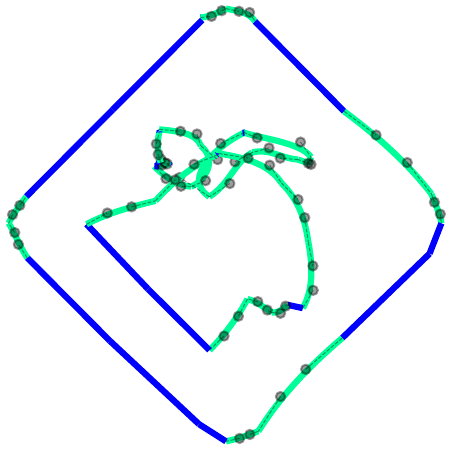} \\
    \end{tabular}

    \caption{
        \textbf{Qualitative comparison of icon reconstructions}:
        Each pair shows ground truth (GT) and our generated glyph.
    }
    \label{fig:icon_reconstruction_comparison}
\end{figure}

Due to the structural differences between icons and typographic glyphs—icons typically contain more paths but fewer commands per path—we train the same model configuration while allowing up to 10 paths per icon and 32 commands per path, setting a maximum of 128 commands per icon. As shown in Figure~\ref{fig:icon_reconstruction_comparison}, the model produces high-quality reconstructions, faithfully recovering icon geometries from their latent representations and demonstrating strong generalization beyond the font domain.

\subsection{Latin Typefaces One-Shot Generation}

In this subsection, we evaluate our model on the task of Latin Typeface Generation. We focus on \emph{cross-reconstruction}, that is, generating glyphs that differ from those provided as input for style encoding.

For our font generative model, DesigNet, we train from a pretrained VAE and increase the model’s capacity to handle the added complexity introduced by cross-reconstruction. For the Latin fonts, the architecture is scaled to 10 encoder and decoder layers with 8 attention heads each, and the feed-forward dimensionality in both the model and latent projections is doubled. To ensure a fair comparison with state-of-the-art models, we limit the maximum sequence length to 64 commands per glyph.

Other methods, such as DeepVecFont-v2, reconstruct a single target glyph by randomly sampling a subset of reference characters. Instead, we adopt a fixed reference set \{\texttt{H}, \texttt{a}, \texttt{m}, \texttt{b}, \texttt{u}, \texttt{r}, \texttt{g}, \texttt{e}\}. This selection follows common practice in type design: these glyphs cover a wide range of structural and geometric variations, including vertical stems, round counters, diagonals, and curves with descenders~\cite{cheng2020designing}. From each reference set, we reconstruct the 52 letters of the Latin alphabet.

We evaluate six configurations: DualVector, the original DeepVecFont-v2 model in both one-shot and few-shot settings, DeepVecFont-v2 augmented with our Self-Refinement modules, and our proposed DesigNet, evaluated both with and without Self-Refinement. For a fair comparison, both DeepVecFont-v2 and DualVector are fine-tuned on our proprietary dataset, starting from their officially released checkpoints.

\begin{table}[ht]
  \centering
  \resizebox{\columnwidth}{!}{
  \begin{tabular}{@{} l *{5}{c} @{}}
    \toprule

    \textbf{Model} & \textbf{IoU} $\uparrow$ & \textbf{L1} $\downarrow$ & \textbf{RE} $\downarrow$& \textbf{Cont. Acc.} $\uparrow$ & \textbf{Align. Acc.} $\uparrow$ \\
    \midrule
        DualVector~\cite{liu2023dualvectorunsupervisedvectorfont}  & 0.564 & 0.137 & - & - & - \\
        DeepVecFont-v2~\cite{deepvecfont2_wang2023} & 0.681 & 0.117 & 13.124 & 0.444 & 0.391 \\

        DeepVecFont-v2 + Self-Ref.                  & 0.675 & 0.120 & 13.269 & \textbf{0.528} & 0.391 \\
        DesigNet w/o Self-Ref.                      & \textbf{0.711} & \textbf{0.106} & \textbf{12.665} & 0.276 & 0.282 \\
        DesigNet                                    & 0.693 & 0.115 & 13.126 & 0.482 & \textbf{0.531}
        \\
        \midrule
        DeepVecFont-v2 10 shots                     & 0.735 & 0.091 & 11.848 & 0.452 & 0.383 \\

    \bottomrule
  \end{tabular}
  }
  \caption{Comparison on our proprietary dataset for Latin typeface generation (52 letters) on the cross-reconstruction task.}
  \label{tab:latin-font-generation}
\end{table}

As shown in Table~\ref{tab:latin-font-generation}, our baseline model outperforms DeepVecFont-v2  in the single-shot setting across the main quantitative metrics (IoU, $\ell_1$, and RE).

Both models benefit from adding the proposed Self-Refinement modules, which consistently improve continuity and alignment accuracy. While these modules may slightly degrade reconstruction metrics (IoU, $\ell_1$, and RE), Fig. \ref{fig:self_refinement_samples} demonstrates that the resulting outputs are visually more appealing. Notably, for DeepVecFont-v2, the Alignment Self-Refinement module does not yield improvements. This behavior is expected, as the use of discrete coordinates inherently enforces alignment whenever points fall within the same discretization bin.

\begin{figure*}[t]
  \centering
  \scriptsize
  \begin{tabular}{@{}c@{\hspace{1em}}c@{}}
    Encoding glyphs & Decoding glyphs \\
    \includegraphics[width=0.20\textwidth]{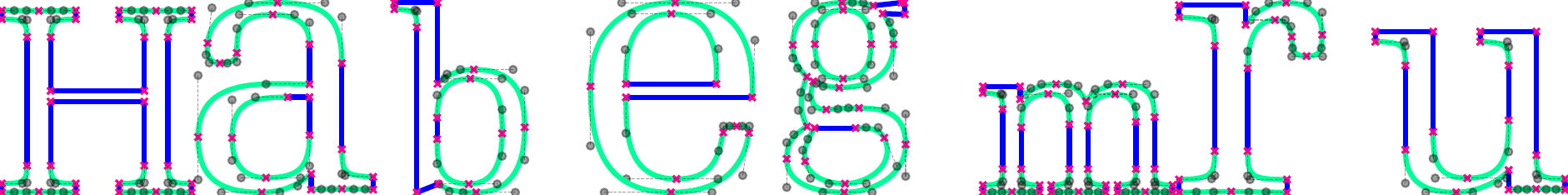} &
    \includegraphics[width=0.80\textwidth]{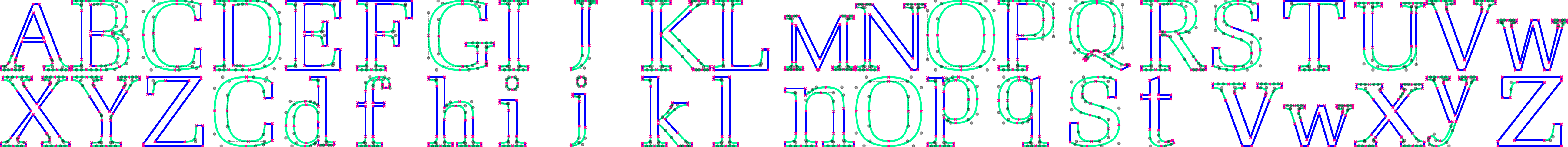} \\
    \multicolumn{2}{c}{\scriptsize GT} \\[5pt]

    \includegraphics[width=0.20\textwidth]{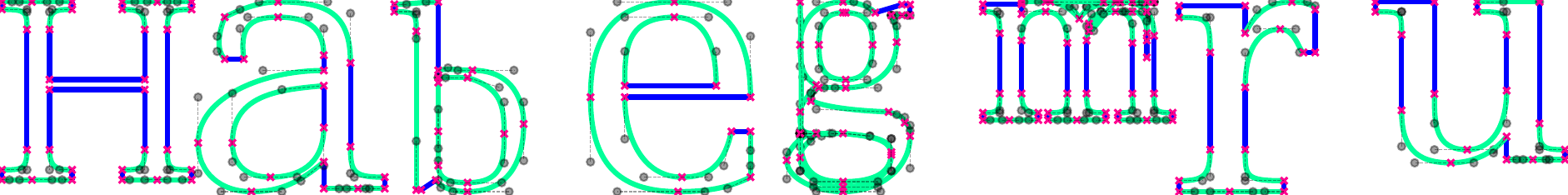} &
    \includegraphics[width=0.80\textwidth]{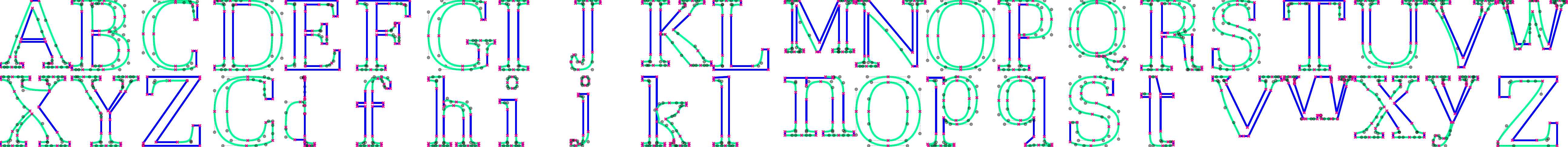} \\
    \multicolumn{2}{c}{\scriptsize DeepVecFont-v2 \cite{deepvecfont2_wang2023}} \\[5pt]

    \includegraphics[width=0.20\textwidth]{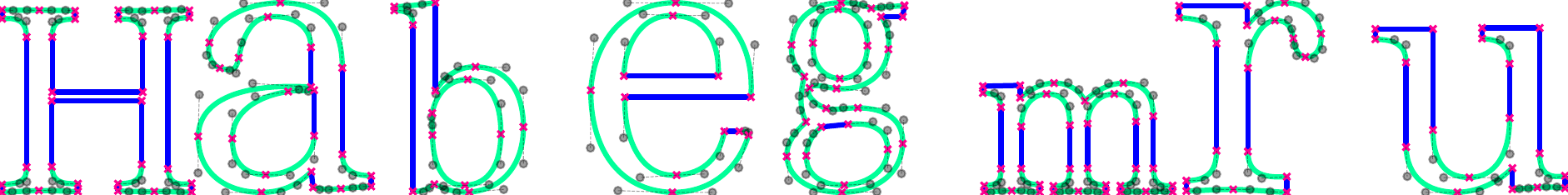} &
    \includegraphics[width=0.80\textwidth]{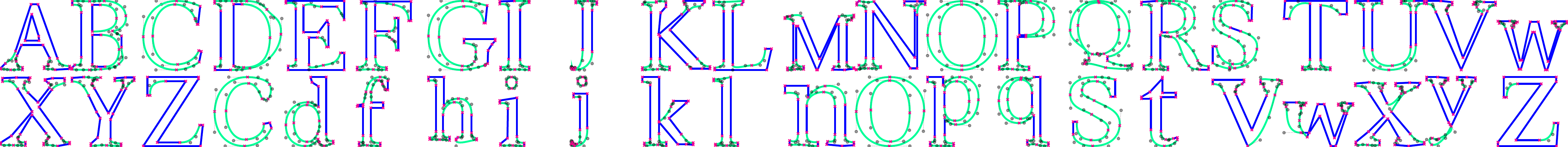} \\
    \multicolumn{2}{c}{\scriptsize Ours} \\
  \end{tabular}

  \caption{Reconstruction quality across encoding (left) and decoding (right) glyph sets. From top to bottom: ground truth (GT), DeepVecFont-v2, and our method after refinement.}
  \label{fig:font_recon_panels}
\end{figure*}

Qualitative comparisons (Fig.~\ref{fig:font_recon_panels}) better illustrate the impact of self-refinement, showing smoother curve transitions, cleaner alignments, and overall sharper glyph structures. Most importantly, these results demonstrate that operating in a continuous argument space, which has long been considered challenging for SVG generation, can yield high-quality reconstructions while maintaining geometric regularity.

Regarding DualVector~\cite{liu2023dualvectorunsupervisedvectorfont}, its objective differs fundamentally from ours. Rather than reproducing the original SVG command structure of the target glyphs, DualVector represents glyphs as unions of learned dual-part primitives (positive and negative closed Bézier paths), which are subsequently combined through boolean operations. The final SVG is obtained via an inference-time refinement procedure that iteratively optimizes control points using differentiable rendering to match a predicted raster image. This process is computationally expensive and may alter topology, path decomposition, and command structure. Consequently, metrics that depend on command-level correspondence or geometric regularity, such as reconstruction error (RE), continuity accuracy, or alignment accuracy, are not directly comparable. For this reason, we restrict the quantitative evaluation of DualVector to image-based metrics (IoU and $\ell_1$), which better reflect its optimization objective and output representation.

In the few-shot setting, DeepVecFont-v2 generates both SVG and raster predictions for each font and selects, among the multiple SVG outputs, the one whose rasterization achieves the highest IoU with its own predicted image. This inference-time selection strategy leverages the fact that, in their dual-branch formulation, raster predictions tend to be more visually faithful and stable; consequently, coherence between the predicted image and its rasterized SVG serves as a proxy for output quality. Such a strategy is not applicable to our approach, which operates exclusively in the SVG domain and does not rely on raster predictions. With 10 shots, DeepVecFont-v2 attains slightly higher scores, but at the cost of substantially increased computation time (approximately $10\times$).

\subsection{Chinese Typefaces One-Shot Generation}
In this subsection, we evaluate our model on Chinese typeface one-shot generation to further assess its generalization capabilities. We train and evaluate on the Chinese font dataset introduced in DeepVecFont-v2~\cite{deepvecfont2_wang2023}. In this setting, each glyph is represented using up to four paths and a maximum of 71 commands.

Due to the smaller size of the dataset, we adopt a reduced architecture consisting of 5 encoder and 5 decoder layers. To maintain comparability with the Latin typeface experiments, we use a reference set of 8 glyphs and reconstruct 52 target glyphs under the cross-reconstruction protocol.

\begin{table}[ht]
  \centering
  \resizebox{\columnwidth}{!}{
  \begin{tabular}{@{} l *{6}{c} @{}}
    \toprule

    \textbf{Model} & \textbf{IoU} $\uparrow$ & \textbf{L1} $\downarrow$ & \textbf{RE} $\downarrow$ & \textbf{Cont. Acc.} $\uparrow$ & \textbf{Align. Acc.} $\uparrow$ \\
    \midrule
        DeepVecFont-v2~\cite{deepvecfont2_wang2023} & 0.380 & \textbf{0.216} & 21.414 & 0.499 & 0.198 \\
        DeepVecFont-v2 + Self-Ref.                  & 0.378 & 0.218 & 21.423 & 0.501 & 0.205  \\
        DesigNet w/o Self-Ref.                      & \textbf{0.397} & 0.219 & \textbf{19.604} & 0.501 & 0.247 \\
        DesigNet                                    & 0.391 & 0.221 & 19.651 & \textbf{0.512} & \textbf{0.351} \\
    \midrule
        DeepVecFont-v2 10 shots                     & 0.417 & 0.199 & 20.211 & 0.498  & 0.202 \\
    \bottomrule
  \end{tabular}
  }
  \caption{Comparison on the Chinese font dataset~\cite{deepvecfont2_wang2023}. Metrics are computed on the cross-reconstruction task.}
  \label{tab:chinese-font-generation}
\end{table}

\section{Conclusions}
\label{sec:conclusions}
We presented DesigNet, a hierarchical Transformer-VAE that operates natively in the SVG domain with continuous commands and designer-oriented controls.

By predicting and enforcing junction continuity ($C^0$, $G^1$, $C^1$) and axis alignment through deterministic self-refinement, the model produces editable outlines that better match professional practice.

On Latin and Chinese benchmarks, DesigNet improves IoU, image $\ell_{1}$, and reconstruction error over the baselines, substantially increasing continuity and alignment accuracy. The resulting SVG paths load easily into commercial software such as FontForge, Glyphs 3, and Adobe Illustrator, which makes downstream editing straightforward.

Despite strong progress, our outputs and those of the current state of the art still fall short of professional standards for style consistency across weight, contrast, slant, and aperture. Operating in absolute coordinates limits the ability to copy or tie repeated structures across glyphs (for example, identical serifs), which hinders exact motif reuse. Promising directions include diffusion or flow-matching decoders and explicit compositionality that assembles glyphs from reusable parts, especially for ideographic scripts such as Chinese.

\bibliographystyle{eg-alpha-doi}
\bibliography{egbib}

\end{document}